\documentclass[conference]{IEEEtran}
\IEEEoverridecommandlockouts
\usepackage{cite}
\usepackage{amsmath,amssymb,amsfonts}
\usepackage{algorithmic}
\usepackage{graphicx}
\usepackage{textcomp}
\usepackage{xcolor}
\usepackage{balance}

\usepackage[ruled,vlined,linesnumbered]{algorithm2e}

\usepackage{booktabs} 
\usepackage{subcaption} 

\usepackage{fancyhdr}
\usepackage{tabularx}
\usepackage{multirow}
\usepackage{multicol}
\usepackage{mdwlist}
\usepackage{indentfirst}
\usepackage{physics}
\usepackage{xspace}
\usepackage{tablefootnote}
\usepackage{footnote}
\usepackage[bottom]{footmisc}
\usepackage{hyperref}
\usepackage{makecell}
\usepackage{subcaption}
\xspaceaddexceptions{]\}}

\newcommand{\method}{\textsc{DaConA}\xspace}

\def\BibTeX{{\rm B\kern-.05em{\sc i\kern-.025em b}\kern-.08em
    T\kern-.1667em\lower.7ex\hbox{E}\kern-.125emX}}

\begin{document}
\bstctlcite{IEEEexample:BSTcontrol}
\title{Data Context Adaptation for Accurate Recommendation with Additional Information}
\author{
\IEEEauthorblockN{Hyunsik Jeon}
\IEEEauthorblockA{Seoul National University\\
Seoul, Republic of Korea\\
jeon185@snu.ac.kr}
\and
\IEEEauthorblockN{Bonhun Koo}
\IEEEauthorblockA{Seoul National University\\
Seoul, Republic of Korea\\
darkgs@snu.ac.kr}
\and
\IEEEauthorblockN{U Kang}
\IEEEauthorblockA{Seoul National University\\
Seoul, Republic of Korea\\
ukang@snu.ac.kr}}

\maketitle

\begin{abstract}
Given a sparse rating matrix and an auxiliary matrix of users or items, how can we accurately predict missing ratings considering different data contexts of entities? 
Many previous studies proved that utilizing the additional information with rating data is helpful to improve the performance.
However, existing methods are limited in that
1) they ignore the fact that data contexts of rating and auxiliary matrices are different,
2) they have restricted capability of expressing independence information of users or items, and 
3) they assume the relation between a user and an item is linear.

We propose \method, a neural network based method for recommendation with a rating matrix and an auxiliary matrix.
\method is designed with the following three main ideas.
First, we propose a data context adaptation layer to extract pertinent features for different data contexts.
Second, \method represents each entity with latent interaction vector and latent independence vector. Unlike previous methods, both of the two vectors are not limited in size.
Lastly, while previous matrix factorization based methods predict missing values through the inner-product of latent vectors, \method learns a non-linear function of them via a neural network.
We show that \method is a generalized algorithm including the standard matrix factorization and the collective matrix factorization as special cases.
Through comprehensive experiments on real-world datasets, we show that \method provides the state-of-the-art accuracy.
\end{abstract}


\section{Introduction}
\label{sec:intro}

Given a sparse rating matrix and an auxiliary matrix of users or items, how can we accurately predict missing ratings considering different data contexts (e.g., an item belongs to both rating and item-genre contexts) of entities?
%
Predicting unseen rating values is a crucial problem in recommendation because users want to be provided an item list that they will give high ratings.

Matrix factorization (MF)~\cite{koren2009mfrecommender, salakhutdinov2008pmf} is a basic yet extensively used method in recommendation due to its simplicity and powerful performance.
Given only a sparse rating matrix $X$, MF derives two low-rank latent matrices $U$ and $V$ that represent user and item features, respectively.
It optimizes $U$ and $V$ to reduce the loss $\|X - U^\intercal V\|_F^2$ for observed ratings, where $\|\cdot\|_F^2$ denotes the Frobenius norm.
Then MF predicts unobserved rating that user $i$ will give to item $j$ as $U_i^\intercal V_j$, where $U_i$ is $i$th column of $U$ and $V_j$ is $j$th column of $V$.
The latent vectors $U_i$ and $V_j$ are trained to represent interaction information.
In the real world, however, there exist independence information of users or items which does not directly interact with other information.
Biased-MF has been proposed to incorporate such information as well.
Biased-MF predicts unobserved rating between user $i$ and item $j$ as $U_i^\intercal V_j + b_i + b_j$, where $b_i \in \mathbb{R}$ and $b_j \in \mathbb{R}$ are bias terms of user $i$ and item $j$, respectively.
%
However, biased-MF has a restricted model capacity since the inner-product yields only a 1-d scalar value.
%
MF also has a limitation that it models only a linear function.
For example, suppose that each dimension of $U_i$ and $V_j$ is trained to represent the degree of each characteristic (e.g., comedy, horror and action in movie genres).
MF multiplies the corresponding degrees of characteristics and adds them to predict the rating. 
In the real world, however, the relationship between a user and an item is not always linear.
A user may dislike a movie thoroughly if the degree of fear is below a certain level, regardless of the degree of other characteristics.
In order to overcome the limitation of linear model, deep-learning approach is introduced to MF~\cite{dziugaite2015neural, he2017neural}.
However, they only utilize a rating matrix although additional information is available in many services.

\begin{table}[t]
	\caption{Comparison of \method and other methods.
	Bold fonts indicate desired settings.
	\method is the only method supporting all the desired properties and providing the richest modeling capability.}
	\centering
	\label{table:salesman}
	\resizebox{8.5cm}{!}{
	\begin{tabular}{c | c c c c}
		\toprule
		\textbf{Method} & \begin{tabular}{@{}c@{}}\textbf{Use} \\ \textbf{additional data}\end{tabular} &\begin{tabular}{@{}c@{}}\textbf{Consider} \\ \textbf{data context} \\ \textbf{difference}\end{tabular} & \begin{tabular}{@{}c@{}}\textbf{Model} \\ \textbf{independence} \\ \textbf{information}\end{tabular} &\textbf{Linearity} \\
		\midrule
		MF~\cite{koren2009mfrecommender, salakhutdinov2008pmf} & X & X & X & Linear \\
		Biased-MF & X & X & Restricted & Linear \\
		NeuMF~\cite{he2017neural} & X & X & \textbf{O} & \textbf{Non-linear} \\
		\midrule
		CMF~\cite{singh2008relational} & \textbf{O} & X & X & Linear \\
		Biased-CMF & \textbf{O} & X & Restricted & Linear \\
		FM~\cite{rendle2010factorization, rendle2012factorization, blondel2016factorization} & \textbf{O} & X & Restricted & Linear \\
		SREPS~\cite{liu2018social} & \textbf{O} & \textbf{O} & X & Linear \\
		HybridCDL~\cite{dong2017hybrid} & \textbf{O} & X & X & \textbf{Non-linear} \\
		\textbf{\method (proposed)} & \textbf{O} & \textbf{O} & \textbf{O} & \textbf{Non-linear}\\
		\bottomrule
	\end{tabular}}
	\vspace{0.2cm}
\end{table}

In recent years, numbers of algorithms have been proposed to use additional information (e.g., social networks~\cite{hao2008sorec, ma2009learning, jamali2010mftrustpropagation, ma2011soreg, yang2013trustmf, tang2016recommendation}, item characteristics~\cite{singh2008relational, li2015deep, dong2017hybrid}, and item synopsis~\cite{wang2011collaborative, wang2015collaborative, kim2016convolutional}) as well as rating data to improve the performance of recommendation, and they have shown that utilizing both rating data and auxiliary data helps improve the accuracy of rating prediction; thus, effective usage of auxiliary data beyond the rating data has become an important issue in recommendation. 
We call them as \textit{data context-aware recommendation}, which is different from \textit{context-aware recommendation systems} (CARS) that consider users' specific situation (e.g., time, place and weather, etc).
Collective Matrix Factorization (CMF)~\cite{singh2008relational} is the most popular method for data context-aware recommendations.
CMF comprises two tied MF models which share a single latent matrix and are trained to minimize the losses for a rating matrix and an additional matrix (details in Section~\ref{sec:prelim}).
CMF is also extended to biased-CMF if it is tied to two biased-MF models.
However, CMF has a limitation that it directly applies a common latent matrix to two MF models without considering different types of data contexts.
For example, assume we have a truster-trustee matrix for auxiliary information, in addition to rating matrix.
CMF then forces user latent vectors to represent users' preferences for both rating and social relationship.
However, sharing the same latent vector in different data contexts may not be proper in the real world; users may mainly regard movie's genre, popularity, and actors to give ratings (rating data context), whereas they consider age, gender, and social status to establish social relationships (social relationship data context).
Furthermore, CMF inherits MF's limitation of predicting the preferences as a linear function and not sufficiently considering independence information.


In this paper, we propose \method (Data Context Adaptation for Accurate Rating Recommendation with Additional Information), an accurate recommendation framework based on deep neural networks using both rating matrix and auxiliary data.
\method is based on the following three main ideas.

\begin{figure}[t]
	\centering
	\hspace{-15px}
	\includegraphics[width=0.42\textwidth]{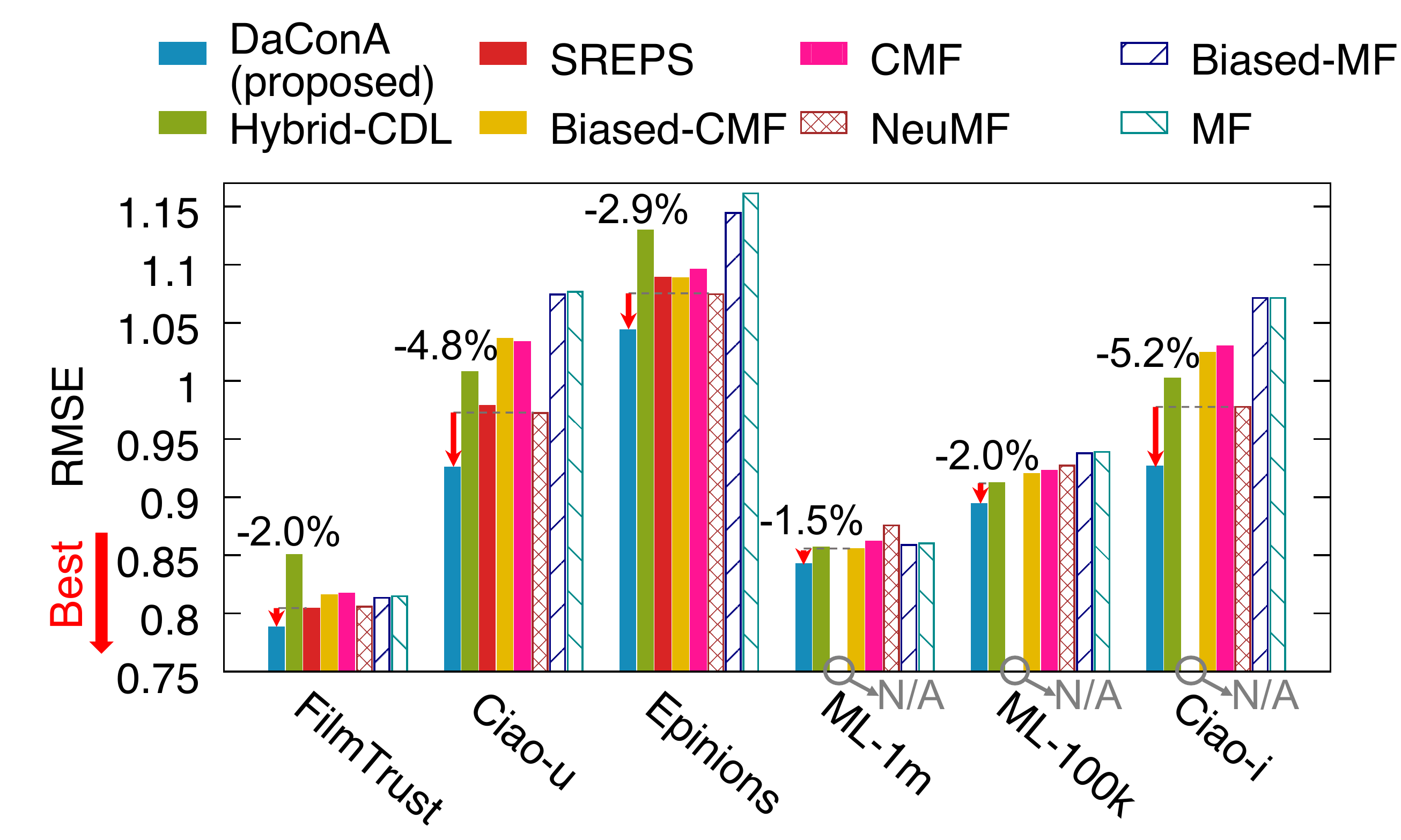}
	\vspace{-5px}
	\caption{
	\method shows the best accuracy (lowest RMSE) for six real-world datasets;
	each percentage indicates the error reduction rate compared to the second best one in each dataset.
	We omit FM in the plot since it shows far worse performance than the other methods.
	SREPS is available only for user-coupled datasets.
	}
	\label{fig:overall}
\end{figure}

\begin{enumerate*}

\item We propose data context adaptation layer in \method to adapt latent interaction vector to different data contexts: rating data context and auxiliary data contexts.
    The data context adaptation layer extracts pertinent features appropriate for different data contexts.

\item \method represents each entity (e.g., a user and an item) with two types of vectors: \textit{latent interaction vector} and \textit{latent independence vector}.
Latent interaction vector is optimized to represent information that is interactive with other entities, whereas latent independence vector is optimized to represent information that is not interactive with other entities.
Unlike biased-MF and biased-CMF which use a scalar bias for each entity, latent independence vector in \method is not limited in size.
%

\item We design a neural network framework to predict non-linear relationship between entities.
We combine two neural networks using an integrated loss function; they are simultaneously trained to minimize the loss.

%

\end{enumerate*}

Table~\ref{table:salesman} compares \method with other algorithms in various perspectives.
\method is the only method that uses additional data, exploits data context difference, models rich independence information, and models non-linear relationship. 

The contributions of \method are as follows:
\begin{itemize}
	\item \textbf{Algorithm.} We propose \method, a method to predict missing values in collective data using neural network. \method learns a non-linear function and latent interaction/independence factors, and adapts the latent interaction factors into different contexts via data context adaptation layer.
	\item \textbf{Generalization.} We show that \method is a generalized algorithm of well-known MF and Collaborative Filtering algorithms, and further provides rich modeling capability.
	\item \textbf{Performance.}
\method provides the best accuracy, up to 5.2\% lower RMSE than the second best method in real-world datasets (see Figure~\ref{fig:overall}).
We also show that our key ideas improve the performance.
\end{itemize}


Table~\ref{table:symbols} lists the symbols used in this paper. 
Our source code and datasets are available at \url{https://datalab.snu.ac.kr/dacona/}. 

\begin{table}[t]
	\caption{Table of frequently-used symbols.}
	\centering
	\label{table:symbols}
		\resizebox{8.0cm}{!}{
		\begin{tabular}{ c | l }
			\toprule
			\textbf{Symbol} &	\textbf{Definition}	\\
			\midrule
			$\mathbb{U}$, $\mathbb{I}$, $\mathbb{C}$ & set of users, items, and context entities\\
			$X \in \mathbb{R}^{|\mathbb{U}| \times |\mathbb{I}|}$	&	rating matrix\\
			$Y \in \mathbb{R}^{|\mathbb{I}| \times |\mathbb{C}|}$ &  additional matrix for item\\
			$\Omega_X$, $\Omega_Y$ & set of indices of observable entries in $X$ and $Y$\\
			\midrule
			$U$, $V$, $C$	&	latent interaction matrices of user, item,\\
			& and additional entity (e.g., trustee and genre)\\
			$U^X$, $V^X$	&	latent independence matrices of user\\
			& and item in rating data context \\
			$V^Y$, $C^Y$	&	latent independence matrices of item\\
			& and additional entity in auxiliary data context \\
			$D^X$, $D^Y$	&	data context adaptation matrices\\
			&  for rating data context and auxiliary data context \\
			$f_X$, $f_Y $ & fully-connected neural networks for $X$ and $Y$\\
			$\theta_{f_X}$, $\theta_{f_Y}$ & parameters of $f_X$ and $f_Y$\\
			\midrule
			$d_c$ & dimension of latent interaction vector\\
			$d_s$ & dimension of latent independence vector\\
			\bottomrule
		\end{tabular}}
\end{table}


\section{Preliminaries: Collective Matrix Factorization}
\label{sec:prelim}
Traditional Matrix Factorization (MF) suffers from a data sparsity problem~\cite{su2009survey};
Collective Matrix Factorization (CMF)~\cite{singh2008relational} has been proposed to solve the problem by using additional data.
While MF decomposes a single matrix into two latent matrices, CMF decomposes both rating matrix and additional matrix into three latent matrices, and minimizes an integrated loss function which is based on \textit{inner-product}.
For a factorization problem of users and items,
either user context matrix or item context matrix is available as an additional matrix.
Assuming the auxiliary matrix is item-coupled,
given a rating matrix $X \in \mathbb{R}^{|\mathbb{U}| \times |\mathbb{I}|}$ and an auxiliary matrix $Y \in \mathbb{R}^{|\mathbb{I}| \times |\mathbb{C}|}$,
rank-$d$ decomposition of these two data matrices yields three latent matrices,
$U \in \mathbb{R}^{d \times |\mathbb{U}|}, V\in \mathbb{R}^{d \times |\mathbb{I}|}$ and $C\in \mathbb{R}^{d \times |\mathbb{C}|}$.
$\mathbb{U}$, $\mathbb{I}$, and $\mathbb{C}$ indicate set of users, items, and auxiliary entities, respectively.
$V$ is shared for the predicted matrices $\hat{X}$ and $\hat{Y}$ as follows:
\begin{equation*}
\footnotesize
\hat{X}_{ij} = U_i^\intercal V_j, \;\; \hat{Y}_{jk} = V_j^\intercal C_k,
\end{equation*}


\noindent where $U_i$, $V_j$, and $C_k$ are $i$th column of $U$, $j$th column of $V$, and $k$th column of $C$, respectively.
$U$, $V$, and $C$ are trained to minimize the following loss function:
\begin{equation*}
\footnotesize
\begin{aligned}
L = \frac{1}{2} \sum_{(i,j) \in \Omega_X} (\hat{X}_{ij}- X_{ij})^2
+ \frac{1}{2} \sum_{(j,k) \in \Omega_Y} (\hat{Y}_{jk} - Y_{jk} )^2 \\
+ \frac{\lambda}{2}(\|U|\|_F^2 + \|V|\|_F^2 + \|C|\|_F^2),
\end{aligned}
\end{equation*}
\noindent where $\Omega_X$ and $\Omega_Y$ are sets of indices of observable entries in $X$ and $Y$, respectively.
$\lambda$ is a regularization parameter to prevent overfitting.
CMF is extended to biased-CMF if bias terms are added.
Biased-CMF predicts ratings and entries of auxiliary matrix as follows:
\begin{equation*}
\hat{X}_{ij} = U_i^\intercal V_j + b_i + b_j, \;\; \hat{Y}_{jk} = V_j^\intercal C_k + \bar{b}_j + \bar{b}_k,
\end{equation*}
where $b_i, b_j, \bar{b}_j, \bar{b}_k \in \mathbb{R}$ are bias terms.
The bias terms represent independence information of entities.

CMF based method~\cite{hao2008sorec} shows better performance than MF~\cite{koren2009mfrecommender, salakhutdinov2008pmf} for real-world datasets. 
However, CMF has limitations that it predicts the relationship between entities by linear functions and does not consider the fact that data context of the rating matrix and that of the additional matrix are different.
Moreover, even biased-CMF is not able to sufficiently consider independence information because the dimensions of all bias terms are restricted: each entity has a scalar bias term.


\begin{figure*}[t]
\centering
\includegraphics[width=0.8\textwidth]{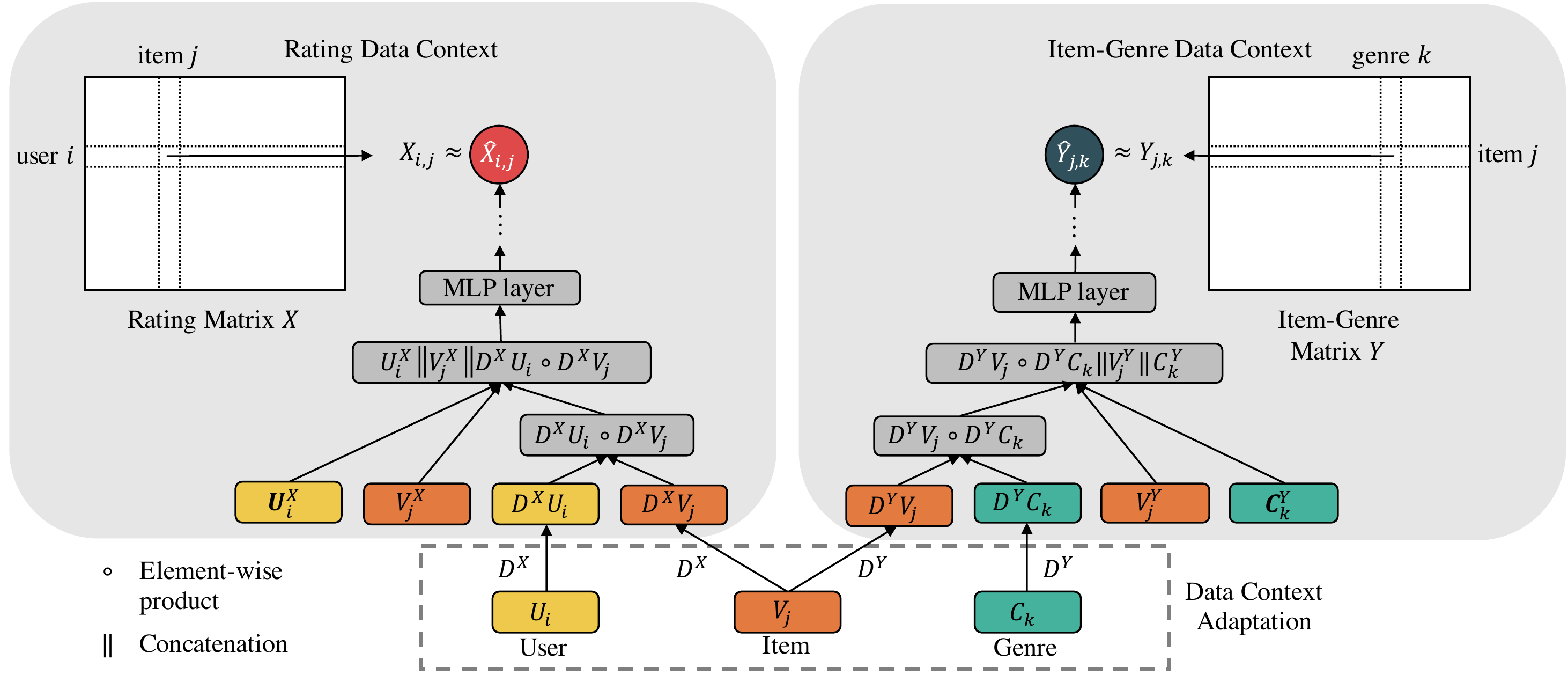}
\caption{
Architecture of \method for item coupled dataset.
}
\label{fig:deepcmf}
\end{figure*}

\section{Proposed Method}
\label{sec:proposed}

We describe our proposed method \method for rating prediction.
After presenting the overview of our method in Section~\ref{subsec:310}, we formulate the objective function of \method in Section~\ref{subsec:320}.
Then we describe three key ideas of \method in Sections~\ref{subsec:330}, ~\ref{subsec:340}, and ~\ref{subsec:350}, respectively.
Afterwards, we explain the training algorithm of \method in Section~\ref{subsec:360}.
Finally, we compare \method with existing methods in Section~\ref{subsec:370} showing that \method is a generalized version of them.
	

\subsection{Overview}
\label{subsec:310}
\method utilizes a rating matrix and an auxiliary matrix to predict unseen ratings.
We design \method based on the following observations:
1) there exist independence factors such as biases as well as interaction factors that affect the relationship between entities in real-world,
2) the information needed for different data contexts is different, and
3) users and items do not interact in a simple linear way.
Based on these observations, the goals of \method are
1) to learn embedding vectors that represent interaction and independence features of entities,
2) to extract appropriate features for each data context from a shared latent factor, and
3) to learn complex non-linear scoring function between entities.
Designing such recommendation model entails the following challenges:

\begin{enumerate}
	\item \textbf{Modeling entities.} How can we model the embedding vector for each entity to represent interaction and independence features?
	\item \textbf{Considering data context difference.} How can we learn the common factor between rating matrix and an auxiliary matrix (e.g., item-genre matrix or truster-trustee matrix) even though they are in different data contexts?
	\item \textbf{Non-linear relationship.} How can we learn a scoring function that grasps non-linear relationship between entities?
\end{enumerate}

We propose the following main ideas to address the challenges:

\begin{enumerate}
	\item \textbf{Latent interaction and independence factors.} We map each entity into two types of latent factors: \textit{latent interaction factor} which captures interaction features between entities, and \textit{latent independence factor} which captures independence features.
	\item \textbf{Data context adaptation.} We adapt the latent interaction vectors to each data context through data context adaptation layer before scoring the relationship between two entities.
	\item \textbf{Non-linear modeling.} Fully-connected neural networks are trained to score between entities using latent independence factors, and latent interaction factors that are adapted to each data context.
Neural networks are capable of seizing non-linear relationship between features because of non-linear activation functions.
\end{enumerate}

Figure~\ref{fig:deepcmf} shows the architecture of \method.
Latent interaction vectors ($U_i$, $V_j$, and $C_k$) are projected into each data context by multiplication with data context adaptation matrices ($D^X$ and $D^Y$).
We perform element-wise product on the projected latent interaction vectors to capture interactions between two entities.
We then concatenate them with latent independence vectors of entities and feed them into fully-connected neural networks.
The networks predict the entries of rating matrix ($\hat{X}_{ij}$) and the entries of auxiliary matrix ($\hat{Y}_{jk}$).
In the following sections, we formulate \method's objective function, describe \method's predictive model in details, and compare \method with MF and CMF showing that they are limited settings of \method.
For simplicity, we assume that the auxiliary matrix $Y$ is an items' auxiliary matrix representing genre data context (containing item and genre) in the remainder of this section; it is trivial to use users' auxiliary matrix such as truster-trustee matrix instead.
Extending \method to utilize multiple additional information, and its experimental results are discussed in Appendix~\ref{subsec:A10}.
%

\subsection{Objective Formulation}
\label{subsec:320}

Let $X \in \mathbb{R}^{|\mathbb{U}| \times |\mathbb{I}|}$ be a rating matrix and $Y \in \mathbb{R}^{|\mathbb{I}| \times |\mathbb{C}|}$ be an items' auxiliary matrix such as item-genre matrix, where $\mathbb{U}, \mathbb{I}$, and $\mathbb{C}$ indicate sets of users, items, and auxiliary entities, respectively.
We then formulate the objective function (to be minimized) of \method as follows:
\begin{equation}\label{eq:totalloss}
\begin{aligned}
L = (1-\alpha)loss_X + \alpha loss_Y,
\end{aligned}
\end{equation}
where
\begin{equation}\label{eq:lossX}
\footnotesize
\begin{aligned}
loss_X &= \frac{1}{2} \sum_{(i, j) \in \Omega_X} (\hat{X}_{ij} - X_{ij})^2 + \frac{\lambda}{2} Reg_X,\\
\end{aligned}
\end{equation}
\begin{equation}\label{eq:lossY}
\footnotesize
\begin{aligned}
loss_Y &= \frac{1}{2} \sum_{(j, k) \in \Omega_Y} (\hat{Y}_{jk} - Y_{jk})^2 + \frac{\lambda}{2} Reg_Y.\\
\end{aligned}
\end{equation}

$\Omega_X$ and $\Omega_Y$ are sets of indices of observable entries in $X$ and $Y$, respectively.
$\hat{X}_{ij}$ and $\hat{Y}_{jk}$ are predicted entries from \method.
$Reg_X$ and $Reg_Y$ are regularization terms controlled by $\lambda$ to prevent overfitting (details in Section~\ref{subsec:350}).
$0 \leq \alpha \leq 1$ is a hyperparameter that controls the balance between $loss_X$ and $loss_Y$.


\subsection{Latent Interaction/Independence Factors}
\label{subsec:330}

\textit{Latent interaction vector} represents the interaction information of entities.
Let $U \in \mathbb{R}^{d_c \times |\mathbb{U}|}$, $V \in \mathbb{R}^{d_c \times |\mathbb{I}|}$, and $C \in \mathbb{R}^{d_c \times |\mathbb{C}|}$ be latent interaction matrices of users, items, and additional entities, respectively, where $d_c$ is the dimension of latent interaction vector.
Then $i$th column $U_i \in \mathbb{R}^{d_c}$ of $U$, $j$th column $V_j \in \mathbb{R}^{d_c}$ of $V$, and $k$th column $C_k \in \mathbb{R}^{d_c}$ of $C$ are user $i$'s, item $j$'s, and additional entity $k$'s latent interaction vector, respectively, which are optimized through training process.
Assume that $\mathbb{I}$ is the set of movies.
If user $i$ likes romance movies and movie $j$ is about romance, the relationship between the two entities has to be strong.
Conversely, if user $i$ likes comic movies but movie $j$ is not comic at all, the relationship between the two entities has to be weak.
Matrix Factorization (MF) captures interaction features of entities by performing inner-product of two latent interaction vectors;
\method generalizes MF by performing element-wise product of two latent interaction vectors, and applies non-linear neural network layers on top of them.
Note that the latent interaction vectors are adapted to each data context as described in Section~\ref{subsec:340}; however, the concept that the latent interaction vectors capture the interaction information does not change.

\textit{Latent independence vector} represents the independence information for each entity $i$, $j$, and $k$.
If item $i$ is of very high quality, then the item will receive high ratings from most users, regardless of their tastes.
Therefore, we need latent independence vectors to represent independence information which latent interaction vectors cannot represent.
Also, an item may receive high ratings from many users in rating data context, while the item does not belong to many genres in the genre data context.
To address such issue, we model each entity to have separate latent independence vector for each data context.
Let $U^X \in \mathbb{R}^{d_s \times |\mathbb{U}|}$ and $V^X \in \mathbb{R}^{d_s \times |\mathbb{I}|}$ be latent independence matrices for users and items, respectively, in rating data context.
Let $V^Y \in \mathbb{R}^{d_s \times |\mathbb{I}|}$ and $C^Y \in \mathbb{R}^{d_s \times |\mathbb{C}|}$ be latent independence matrices for items and genres, respectively, in item-genre data context.
Then $U^X_i \in \mathbb{R}^{d_s}$ and $V^X_j \in \mathbb{R}^{d_s}$ are user $i$'s and item $j$'s latent independence vectors respectively in rating data context.
Similarly, $V^Y_j \in \mathbb{R}^{d_s}$ and $C^Y_k \in \mathbb{R}^{d_s}$ are item $j$'s and genre $k$'s latent independence vectors respectively in item-genre data context. 
All of the latent independence vectors are optimized through training process. 
Note that the dimension of latent independence vector of \method is not limited in size, unlike biased-MF and biased-CMF which is limited to a scalar bias (1-d value).

Formally, the predictive model of \method with latent interaction vectors and latent independence vectors is defined as follows:

\vspace{-0.3cm}

\begin{equation}\label{eq:330}
\begin{aligned}
\hat{X}_{ij} &=f_X(T_X(U_i) \circ T_X(V_j), U^X_i, V^X_j),\\
\hat{Y}_{jk} &=f_Y(T_Y(V_j) \circ T_Y(C_k), V^Y_j, C^Y_k),
\end{aligned}
\end{equation}
where $T_X$ and $T_Y$ denote data context adaptation functions for $X$ and $Y$, respectively (details in Section~\ref{subsec:340}),
$f_X$ and $f_Y$ denote neural networks to predict entries in $X$ and $Y$, respectively (details in Section~\ref{subsec:350}), and
symbol $\circ$ denotes element-wise product operation.
We combine two transferred latent interaction vectors (e.g., $T_X(U_i)$ and $T_X(V_j)$) via element-wise product to learn correlated information.

\subsection{Context Adaptation}
\label{subsec:340}
To capture features appropriate for each data context from latent interaction vectors, we need to learn data context adaptation functions ($T_X$ and $T_Y$) which are in Eq.~(\ref{eq:330}).
We propose a data context adaptation layer which projects the latent interaction vectors into each data context with learnable projection matrices;
the context adaptation layer could be replaced with any projection function such as neural networks.
Let $D^X \in \mathbb{R}^{d'_c \times d_c}$ be a projection matrix for rating data context and $D^Y \in \mathbb{R}^{d''_c \times d_c}$ be a projection matrix for item-genre data context.
Then latent interaction vectors for user $i$ and item $j$ are adapted into rating data context as $D^X U_i$ and $D^X V_j$, respectively.
Here, we apply a single projection matrix $D_X$ for both user and item latent interaction vectors since we want to learn a general mapping to the rating data context from the two interaction vectors, not a separate mapping from each entity;
we experimentally show the effectiveness of sharing the same projection matrix for a data context in Section~\ref{subsec:440}.
Similarly, we adapt latent interaction vectors of item $j$ and genre $k$ to item-genre data context as $D^Y V_j$ and $D^Y C_k$, respectively.
Then Eq.~(\ref{eq:330}) is changed to the following.
\begin{equation}\label{eq:340}
\begin{aligned}
\hat{X}_{ij} &=f_X(D^X U_i \circ D^X V_j, U^X_i, V^X_j),\\
\hat{Y}_{jk} &=f_Y(D^Y V_j \circ D^Y C_k, V^Y_j, C^Y_k).
\end{aligned}
\end{equation}


\subsection{Non-linear Modeling}
\label{subsec:350}
We introduce fully-connected neural networks $f_X$ and $f_Y$ to deal with non-linear relationships between features.
We combine the input features of the networks by concatenating them as follows:
\begin{equation}\label{eq:350}
\footnotesize
\hat{X}_{ij} = f_X
(
\begin{bmatrix}
	D^X U_i \circ D^X V_j \\ U^X_i \\ V^X_j
\end{bmatrix}
),
\hat{Y}_{jk} = f_Y
(
\begin{bmatrix}
	D^Y V_j \circ D^Y C_k \\ V^Y_j \\ C^Y_k \\
\end{bmatrix}
),
\end{equation}
\noindent
where $f_X : \mathbb{R}^{d'_c + 2d_s} \rightarrow \mathbb{R}$ and $f_Y : \mathbb{R}^{d''_c + 2d_s} \rightarrow \mathbb{R}$ are fully-connected neural networks for rating data context and auxiliary data context, respectively.
The square bracket [] denotes concatenation of vectors.
We apply rectifier (ReLU), sigmoid, and hyperbolic tangent (Tanh) as the activation functions in $f_X$ and $f_Y$ from which we observe the followings:
1) although ReLU shows good performance in predicting rating, the variation in the process of convergence is large,
2) sigmoid shows little variation in convergence, but does not show good performance in predicting ratings and the speed of convergence is too slow, and
3) Tanh shows the fastest convergence, little variation, and the best performance.
As a result, we use Tanh as an activation function in \method for experiments.

We use $L2$-regularization to prevent overfitting.
Then $Reg_X$ and $Reg_Y$ respectively in Eq.(\ref{eq:lossX}) and Eq.(\ref{eq:lossY}) are formulated as follows:
\begin{equation}\label{eq:regx}
\footnotesize
\begin{aligned}
Reg_X &= \sum_{i\in\mathbb{U}}(\|U^X_i\|^2 + \|D^XU_i\|^2)\\
&+ \sum_{j\in\mathbb{I}}(\|V^X_j\|^2 + \|D^XV_j\|^2) + \sum\|\theta_{f_X}\|^2,
\end{aligned}
\end{equation}
\begin{equation}\label{eq:regy}
\footnotesize
\begin{aligned}
Reg_Y &= \sum_{j\in\mathbb{I}}(\|V^Y_j\|^2 + \|D^YV_j\|^2)\\
&+ \sum_{k\in\mathbb{C}}(\|C^Y_k\|^2 + \|D^YC_k\|^2) + \sum\|\theta_{f_Y}\|^2,
\end{aligned}
\end{equation}
where $\theta_{f_X}$ and $\theta_{f_Y}$ denote the set of parameters of $f_X$ and $f_Y$, respectively.

\subsection{Training \method}
\label{subsec:360}
Algorithm~\ref{al:optimize} shows how \method predicts the values in $X$ and $Y$, and optimizes the parameters.
We first initialize all parameters via Xavier normalization~\cite{glorot2010understanding} (line~\ref{al:1}).
%
Then we draw samples from $\Omega_X$ (line~\ref{al:3}) and $\Omega_Y$ (line~\ref{al:10}), respectively, and learn parameters for $loss_X$ (lines~\ref{al:4}-\ref{al:8}) and $loss_Y$ (lines~\ref{al:11}-\ref{al:15}) 
alternately.
Specifically, we iteratively optimize a parameter block while fixing the remaining parameter blocks (lines~\ref{al:8},~\ref{al:15}); the derivative of each parameter is calculated with back-propagation.
%
We repeat the update procedure until the validation error converges (line 2).
We adopt Adaptive Moment Estimation (\textit{Adam})~\cite{kingma2014adam} which is a first-order gradient-based optimizer for updating all parameters.
The hyperparameters ($d_c$, $d_s$, $l$, $\lambda$, and $\alpha$) are tuned by grid-search through validation tests.
The optimal settings of hyperparameters are presented in Section~\ref{subsec:410}.

\begin{algorithm}[t]
\small
\algsetup{linenosize=\small}
\DontPrintSemicolon
\SetAlgoLined
\caption{Training \method 
}\label{al:optimize}
\SetKwInOut{Input}{Input}\SetKwInOut{Output}{Output}
\Input{Rating matrix $X \in \mathbb{R}^{|\mathbb{U}| \times |\mathbb{I}|}$,
auxiliary matrix $Y \in \mathbb{R}^{|\mathbb{I}| \times |\mathbb{C}|}$,
set of indices of observable entries $\Omega_X$ and $\Omega_Y$,
balance parameter $\alpha$, and
regularizer $\lambda$}
\Output{latent interaction matrices $U$, $V$, and $C$,
latent independence matrices $U^X$, $V^X$, $V^Y$, and $C^Y$, and
parameters of neural networks $\theta_{f_X}$, and $\theta_{f_Y}$}
\BlankLine
Initialize $U$, $V$, $C$, $U^X$, $V^X$, $V^Y$, $C^Y$, $\theta_{f_X}$, and $\theta_{f_Y}$; \label{al:1}\\
\While{stopping condition is not met}{	\label{al:2}
    \For{$(i,j) \in \Omega_X$}{	\label{al:3}
		$\hat{X}_{ij} \leftarrow$ Eq.~(\ref{eq:350}); \label{al:4}\\
		$Reg_X \leftarrow$ Eq.~(\ref{eq:regx}); \label{al:5}\\
		$loss_X \leftarrow$ Eq.~(\ref{eq:lossX}) with $\hat{X}_{ij}$, $Reg_X$, and $\lambda$; \label{al:6}\\
		$loss \leftarrow (1-\alpha)loss_X$; \label{al:7}\\
		alternately update $U$, $V$, $U^X$, $V^X$, and $\theta_{f_X}$ to minimize $loss$ while fixing the others; \label{al:8}
	}
	\For{$(j,k) \in \Omega_Y$}{	\label{al:10}
		$\hat{Y}_{jk} \leftarrow$ Eq.~(\ref{eq:350}); \label{al:11}\\
		$Reg_Y \leftarrow$ Eq.~(\ref{eq:regy}); \label{al:12}\\
		$loss_Y \leftarrow$ Eq.~(\ref{eq:lossY}) with $\hat{Y}_{jk}$, $Reg_Y$, and $\lambda$; \label{al:13}\\
		$loss \leftarrow \alpha loss_Y$; \label{al:14}\\
		alternately update $V$, $C$, $V^Y$, $C^Y$, and $\theta_{f_Y}$ to minimize $loss$ while fixing the others; \label{al:15}
	}
}
\end{algorithm}

\subsection{Generality of \method}
\label{subsec:370}
We show that \method is a generalized algorithm of several existing methods. 
\begin{itemize}
	\item \textbf{MF}~\cite{koren2009mfrecommender, salakhutdinov2008pmf}.
	\method with the following parameters equals to MF:
	1) setting $\alpha = 0$,
	2) setting $d'_c = d_c$ and $D^X$ to identity matrix,
	3) setting $d_s = 0$, which means there exist no latent independence vectors,
	4) setting the number of layers of $f_X$ to 1 and all weights of the layer to 1, which means the network merely sums all input values,
	and 5) removing activation function, which means that the model captures only linear features.

	\item \textbf{Biased-MF}.
\begin{savenotes}
\begin{table*}[t]
	\centering
	\caption{Dataset statistics.}
	\label{table:datasetstat}
	\resizebox{17cm}{!}{
	\begin{tabular}{r r || r r r r || r r r r}
	\toprule
	&& \multicolumn{4}{c}{\textbf{Rating matrix}} & \multicolumn{4}{c}{\textbf{Auxiliary matrix}}\\
	Dataset name & Auxiliary data type & \# observed ratings & \# users & \# items & density & \# observed entries & \# rows & \# columns & density\\
	\midrule
		Epinions\footnote{\url{http://www.trustlet.org/downloaded_epinions.html}}&user information&658,621&44,434&139,374&0.01\%&487,183&44,434&49,288&0.02\%\\
		Ciao-u\footnote{\label{ciao}\url{https://www.librec.net/datasets/CiaoDVD.zip}}&user information&72,665&18,133&16,121&0.02\%&40,133&18,133&4,299&0.05\%\\
		FilmTrust\footnote{\url{https://www.librec.net/datasets/filmtrust.zip}}&user information&35,093&1,461&2,067&1.16\%&1,853&1,461&732&0.17\%\\
		\midrule
		ML-1m\footnote{\url{https://grouplens.org/datasets/movielens/1m}}&item information&1,000,209&6,040&3,883&4.26\%&73,777&3,883&19&100\%\\
		ML-100k\footnote{\url{https://grouplens.org/datasets/movielens/100k}}&item information&100,000&943&1,682&6.30\%&31,958&1,682&19&100\%\\
		Ciao-i\textsuperscript{\ref{ciao}}&item information&72,655&18,133&16,121&0.02\%&274,057&16,121&17&100\%\\
	\bottomrule
	\end{tabular}}
\end{table*}
\end{savenotes}
	MF is extended to biased-MF if bias term of each entity is added in the loss function.
	Therefore, if we set $d_s = 1$ (at step 3) in the setting of MF above, \method becomes biased-MF.

	\item \textbf{CMF}~\cite{singh2008relational}.
	CMF is composed of two MF models and it has the following settings of \method:
	1) setting $d''_c = d'_c = d_c$, and $D^X$ and $D^Y$ to identity matrix,
	2) setting $d_s = 0$,
	3) setting the number of layers of $f_X$ and $f_Y$ to 1 and all weights of $f_X$ and $f_Y$ as 1,
	and 4) removing activation function.
	
	\item \textbf{Biased-CMF}.
	CMF is extended to biased-CMF by adding bias term of each entity in the loss function.
	Thus, if we set $d_s = 1$ (at step 2) in the setting of CMF above, \method becomes biased-CMF.
\end{itemize}



\section{Experiments}
\label{sec:experiments}

We perform experiments to answer the following questions.
\begin{itemize*}
		\item \textbf{Q1. (Overall performance)}
		How better is \method compared to competitors?
		(Section \ref{subsec:420})
		\item \textbf{Q2. (Effects of interaction and independence factors})
		How do dimensions of \textit{latent interaction and independence vectors} affect the performance?
		(Section \ref{subsec:430})
		\item \textbf{Q3. (Effects of data context adaptation)}
		How does data context adaptation layer affect the performance?
		(Section \ref{subsec:440})
		\item \textbf{Q4. (Neural networks)}
		Do deeper structures yield better performance?
		Is activation function helpful to improve the performance?
		(Section \ref{subsec:450})
\end{itemize*}

\subsection{Experimental Setup}
\label{subsec:410}

\textbf{Metrics.}
We use Root Mean Square Error (RMSE) and Mean Absolute Error (MAE) as evaluation metrics. 

\begin{center}
RMSE  = $\sqrt{\frac{\sum_{u} \sum_{i}(\hat{X}_{ui}-X_{ui})^2}{|test\;ratings|}}$,
MAE  = $\frac{\sum_{u} \sum_{i}|\hat{X}_{ui}-X_{ui}|}{|test\;ratings|}$. \\
\end{center}
Smaller RMSE and MAE indicate better performance since it means the predicted value is closer to the real value.

\textbf{Datasets.}
We use real-world datasets Epinions, Ciao, FilmTrust, MovieLens-1m (ML-1m), and MovieLens-100k (ML-100k) summarized in Table~\ref{table:datasetstat}.
We call Ciao dataset as Ciao-i when using items' auxiliary data, and as Ciao-u when using users' auxiliary data.
Epinions, Ciao-u, and FilmTrust provide users' auxiliary information (user-user trust relationship), while ML-1m, ML-100k, and Ciao-i provide items' auxiliary information (movie-genre).
For each dataset, 80\% of rating data are randomly selected for training and the rest are used for test.
98\% of training set are used for training models, and the rest are used for validation in training process.
All of the auxiliary data are used only for training since our goal is not to predict entries in the auxiliary matrix.

\textbf{Competitors.}
We compare \method to the following state-of-the-art algorithms.
Comparisons of \method and the competitors are summarized in Table~\ref{table:salesman}.
\vspace{-0.1cm}
\begin{itemize*}
	\item MF~\cite{koren2009mfrecommender, salakhutdinov2008pmf}. MF factorizes a rating matrix into user and item factors which are combined via inner-product.
	\item Biased-MF. 1-dimensional bias variables are introduced to MF to model independence factors of entities.
	\item NeuMF~\cite{he2017neural}. NeuMF uses a neural network to capture non-linear relationship between entities.
	\item CMF~\cite{singh2008relational}. CMF is composed of two MFs with an integrated objective function; it learns three low rank matrices by factorizing both rating matrix and auxiliary matrix simultaneously.
	\item Biased-CMF. 1-dimensional bias variables for entities are introduced to CMF.
	\item FM~\cite{rendle2010factorization, rendle2012factorization, blondel2016factorization}. FM factorizes a rating matrix to user, item, and auxiliary entity factors. We use a 2-way FM which captures all single and pairwise interactions. 
	\item SREPS~\cite{liu2018social}. SREPS is composed of matrix factorization model, recommendation network model, and social network model while sharing user latent matrix between the models through separate projections for each model.
	\item Hybrid-CDL~\cite{dong2017hybrid}. Hybrid-CDL consists of two Additional Stacked Denoising Autoencoders (aSDAE), one is for users and the other for items. Each aSDAE encodes rating vector of an entity and auxiliary data vector of the entity to a latent vector. Since we are given one auxiliary matrix with rating matrix, we adopt aSDAE for entities (e.g., item) \emph{with} additional information and adopt SDAE (Stacked Denoising Autoencoders) for entities (e.g., user) \emph{without} additional information.
\end{itemize*}

\begin{table*}[t]
\caption{
\method provides the best accuracy in rating prediction. 
Bold text and * indicate the lowest and the second lowest errors, respectively.
The last column denotes the error reduction rate of \method compared to the second best one.
}
\centering
\label{table:performance}
	\resizebox{17cm}{!}{
	\begin{tabular}{c c || c c c c c c c c c}
		\toprule
		\textbf{Datasets} & \textbf{Metrics} & \textbf{MF} & \textbf{Biased-MF} & \textbf{NeuMF} & \textbf{CMF} & \textbf{Biased-CMF} & \textbf{FM} & \textbf{SREPS} & \textbf{Hybrid-CDL} & \textbf{\method (proposed)}\\
		\midrule
		Epinions		&	RMSE	& 1.1612 & 1.1443 & 1.0746* & 1.0956 & 1.0883 & 1.1557 & 1.0887 & 1.1293 & \textbf{1.0433} (-2.9\%)\\
		(user-coupled)	&	MAE		& 0.8965 & 0.8845 & 0.8270* 	& 0.8513 & 0.8423 & 0.8977 & 0.8462 & 0.8863 & \textbf{0.7996} (-3.3\%)\\
		\midrule
		Ciao-u			&	RMSE	& 1.0765 & 1.0742 &	0.9724* & 1.0568 & 1.0451 & 1.5734 & 0.9786 & 1.0077 & \textbf{0.9255} (-4.8\%)\\
		(user-coupled)	&	MAE		& 0.8349 & 0.8352 &	0.7813 &	 0.8195 & 0.8127 & 1.2018 & 0.7692* & 0.7851 & \textbf{0.6894} (-10.4\%)\\
		\midrule
		FilmTrust		&	RMSE	& 0.8148 & 0.8133 & 0.8058 & 0.8171 & 0.8153 & 0.9053 & 0.8040* & 0.8504 & \textbf{0.7882} (-2.0\%)\\
		(user-coupled)	&	MAE		& 0.6325 & 0.6334 & 0.6240 & 0.6320 & 0.6270 & 0.6769 & 0.6168* & 0.6709 & \textbf{0.6035} (-2.2\%)\\
		\midrule
		ML-1m			&	RMSE	& 0.8601 & 0.8590 & 0.8758 & 0.8561 & 0.8553* & 0.8843 & N/A & 0.8569 & \textbf{0.8422}	(-1.5\%)	\\
		(item-coupled)	&	MAE		& 0.6723 & 0.6715 & 0.6921 & 0.6701 & 0.6691* & 0.6861 & N/A & 0.6715 & \textbf{0.6597} (-1.4\%)	\\
		\midrule
		ML-100k			&	RMSE	& 0.9389 & 0.9378 & 0.9273 & 0.9227	& 0.9200 & 0.9286 & N/A & 0.9118* & \textbf{0.8938} (-2.0\%)	\\
		(item-coupled)	&	MAE		& 0.7415 & 0.7396 & 0.7248 & 0.7234	& 0.7210 & 0.7330 & N/A & 0.7156* & \textbf{0.6988} (-2.3\%)	\\
		\midrule
		Ciao-i			&	RMSE	& 1.0713 & 1.0672 & 0.9777* & 1.0296 & 1.0238 & 1.3387 & N/A & 1.0019 & \textbf{0.9264} (-5.2\%)	\\
		(item-coupled)	&	MAE		& 0.8330 & 0.8293 & 0.7778* & 0.8133 & 0.8130 & 1.0478 & N/A & 0.7830 & \textbf{0.7088} (-8.9\%)\\
		\bottomrule
	\end{tabular}}
\end{table*}

\begin{table}[t]
\caption{Hyperparameters for each dataset. There are no hyperparameters of SREPS for ML-1m, ML-100k, and Ciao-i since SREPS is not available for item-coupled datasets.}
\centering
\label{table:hyperparameters}
	\resizebox{8cm}{!}{
	\begin{tabular}{@{} c @{} c @{} || @{} c @{} c @{} c @{} c @{} c @{} c @{}}
		\toprule
		\multirow{2}{*}{\textbf{Method}} & \textbf{Hyper-} & \multirow{2}{*}{\textbf{Epinions/}} & \multirow{2}{*}{\textbf{Ciao-u/}} &
		\multirow{2}{*}{\textbf{FilmTrust/}} & \multirow{2}{*}{\textbf{ML-1m/}} & \multirow{2}{*}{\textbf{ML-100k/}} & \multirow{2}{*}{\textbf{Ciao-i}}\\
		&\textbf{parameters}&&&&&&\\
		\midrule
		MF, & $l$ & 1e-3 & 1e-3 & 1e-3 & 1e-3 & 1e-4 & 1e-4\\
		Biased-MF& $\lambda$ & 1e-5 & 1e-4 & 1e-3 & 1e-5 & 1e-4 & 1e-4\\
		\midrule
		\multirow{2}{*}{NeuMF} & $l$ & 1e-3 & 1e-4 & 1e-3 & 1e-3 & 1e-3 & 1e-3\\
		& $\lambda$ & 1e-5 & 1e-5 & 1e-4 & 1e-5 & 1e-5 & 1e-4\\
		\midrule
		CMF, & $l$ & 1e-3 & 1e-4 & 1e-3 & 1e-4 & 5e-4 & 1e-4\\
		Biased-CMF& $\lambda$ & 1e-5 & 1e-4 & 1e-4 & 1e-5 & 1e-5 & 1e-5\\
		\midrule
		FM & $l$ & 1e-3 & 1e-3 & 1e-3 & 1e-3 & 1e-3 & 1e-3\\
		\midrule
		\multirow{4}{*}{SREPS} & $l$ & 1e-3 & 1e-3 & 1e-3 & N/A & N/A & N/A\\
		& $\lambda$ & 5e-6 & 1e-5 & 1e-4 & N/A & N/A & N/A\\
		& $\alpha$ & 0.4 & 0.3 & 0.2 & N/A & N/A & N/A\\
		& $\beta$ & 0.1 & 0.2 & 0.1 & N/A & N/A & N/A\\
		\midrule
		\multirow{4}{*}{Hybrid-CDL} & $l$ & 1e-3 & 1e-3 & 1e-3 & 1e-4 & 1e-3 & 1e-3\\
		& $\lambda$ & 1e-2 & 1e-2 & 1e-4 & 1e-3 & 1e-3 & 1e-3\\
		& $\alpha_1$ & 0.8 & 0.2 & 0.5 & 0.2 & 0.2 & 0.2\\
		& $\alpha_2$ & 0.2 & 0.8 & 0.2 & 0.2 & 0.5 & 0.8\\
		\midrule
		& $l$ & 1e-3 & 1e-4 & 1e-3 & 1e-3 & 1e-3 & 1e-3\\
		\method & $\lambda$ & 1e-5 & 1e-5 & 1e-5 & 1e-5 & 1e-5 & 1e-5\\
		(proposed)& $\alpha$ & 0.4 & 0.8 & 0.9 & 0.2 & 0.9 & 0.6\\
		& $d_s$ & 13 & 14 & 11 & 7 & 10 & 11\\
		\bottomrule
	\end{tabular}}
\end{table}

\textbf{Hyperparameters.}
We find the optimal hyperparameters via grid search for all the competitors and \method.
The optimal hyperparameters for each method are shown in Table~\ref{table:hyperparameters}.
We set $d_c = d'_c = d''_c$ in all experiments for simplicity.
We use \textit{tower structure} for $f_X$ and $f_Y$ which is widely used for fully-connected networks~\cite{he2017neural, covington2016deep}.
In the tower structure, dimension of a layer is half of that of the previous layer.
We set the structures of $f_X$ and $f_Y$ to (40, 20, 10) for all experiments.
For fair comparisons, we set the dimensions of predictive factors of all methods to 10; predictive factor indicates the factor that directly decides the ratings.
For instances, predictive factors of each method are as follows: user/item latent vectors of MF, CMF, FM and SREPS, the last hidden layer of NeuMF and \method, and the encoded latent factors of hybrid-CDL.
For NeuMF, we assign 5-dimension of the predictive factor to GMF and the other 5-dimension of that to MLP;
then we use tower structure for the MLP and set $\alpha$ to 0.5 as in~\cite{he2017neural}.
For hybrid-CDL, we adopt [500, 10, 500]-structured hidden layers to aSDAE/SDAE; we set the first dimension of the hidden layers to 500 since it gives the best results among \{1000, 500, 250, 100\} for all experiments.
For biased-MF and biased-CMF, we set the dimensions of user/item latent vectors to 8 since two 1-dimensional bias variables are also used for prediction.

\subsection{Performance Comparison (Q1)}
\label{subsec:420}
We measure rating prediction errors of \method and the competitors in six real-world datasets.
Table~\ref{table:performance} shows RMSEs and MAEs of \method and those of the competitors;
SREPS is only available in user-coupled datasets since it utilizes the auxiliary matrix through network embedding.
Note that
\method consistently outperforms competitors in rating prediction for all six real-world datasets.
Among the competitors, models that use additional information (e.g., CMF) usually perform better than models using only rating information (e.g., MF), suggesting the importance of utilizing the additional information.
Models considering the latent independence factor (e.g., biased-MF and biased-CMF) show better performances than models without the latent independence factor (e.g., MF and CMF) on all datasets, suggesting the importance of considering the independence information.
SREPS outperforms CMF on all user-coupled datasets, suggesting the effectiveness of sharing embedding vectors for different data contexts via an indirect way such as projection.
However, SREPS does not share the projection matrices between the entities in a data context; \method shares projection matrix between entities in a data context which improves the accuracy significantly (details in Section~\ref{subsec:440}).
NeuMF outperforms biased-MF, suggesting the effectiveness of non-linear modeling via neural networks.
However, NeuMF captures non-linear features only from latent independence factors but not from latent interaction factors; \method captures non-linear features from not only latent independence factors but also latent interaction factors since interaction factors also affect the relationship between entities non-linearly.
Hybrid-CDL does not perform well for user-coupled datasets which include extremely sparse auxiliary matrices since it directly feeds sparse auxiliary vectors (e.g., $Y_{j,:}$) into the model.
A supplementary experiment is in Appendix~\ref{subsec:A10} where we compare the performance of \method and hybrid-CDL given two additional information (item-genre matrix and user-trustee matrix).

\subsection{Effects of Interaction and Independence Factors (Q2)}
\label{subsec:430}
\begin{figure}[t]
	\centering
	\hspace{-10px}
	\includegraphics[width=0.35\textwidth]{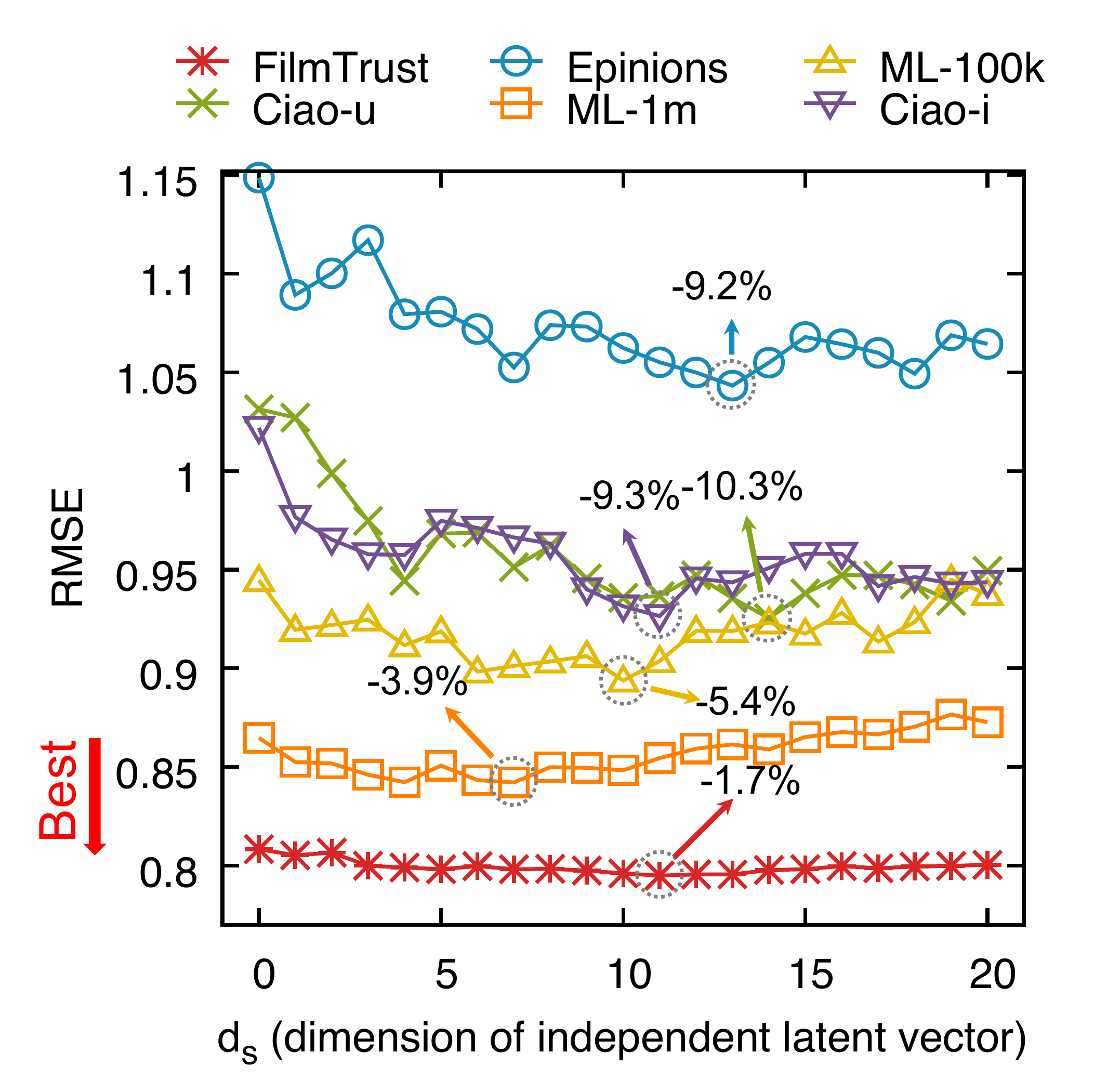}
	\vspace{-5px}
	\caption{
Considering both interaction and independence information is crucial.
The plot shows change of RMSE varying the dimension $d_s$ of latent independence vector.
Each percentage indicates the reduction rate of the best RMSE compared to the worst RMSE in each dataset.
Note that the best RMSE is achieved when $d_s$ is not too small or too large.
	}
	\label{fig:ds}
\end{figure}

We empirically verify the effectiveness of balancing the dimensions of latent interaction and independence factors on real-world datasets in Figure~\ref{fig:ds}.
We measure RMSE of \method while varying the dimension $d_s$ of latent independence vector.
The dimension $d_c$ of latent interaction vector is determined to be ($40 - 2d_s$) to maintain the network structure, and the other hyperparameters are fixed as optimal settings which are reported in Table~\ref{table:hyperparameters}.
%
%
Increasing $d_s$ enlarges the model capacity to embed independence information, while it reduces the capacity to embed interaction information.
We train and test the model for each setting ten times and average them.
RMSEs are the lowest when $d_s$ is set to the followings: (11 in FilmTrust), (14 in Ciao-u), (13 in Epinions), (7 in ML-1m), (10 in ML-100k), and (11 in Ciao-i); conversely, RMSEs are the highest when $d_s$ is set to the followings: (0 in FilmTrust, Ciao-u, Epinions, and Ciao-i), and (19 in ML-1m and ML-100k).
The results show that it is important to balance the interaction and independence information rather than extreme consideration for one of them;
note that more performance improvements are observed for larger and sparser data (Epinions, Ciao-u, and Ciao-i).

\subsection{Effects of Data Context Adaptation (Q3)}
\label{subsec:440}
\begin{figure}[t]
	\centering
	\hspace{-15px}
	\includegraphics[width=0.42\textwidth]{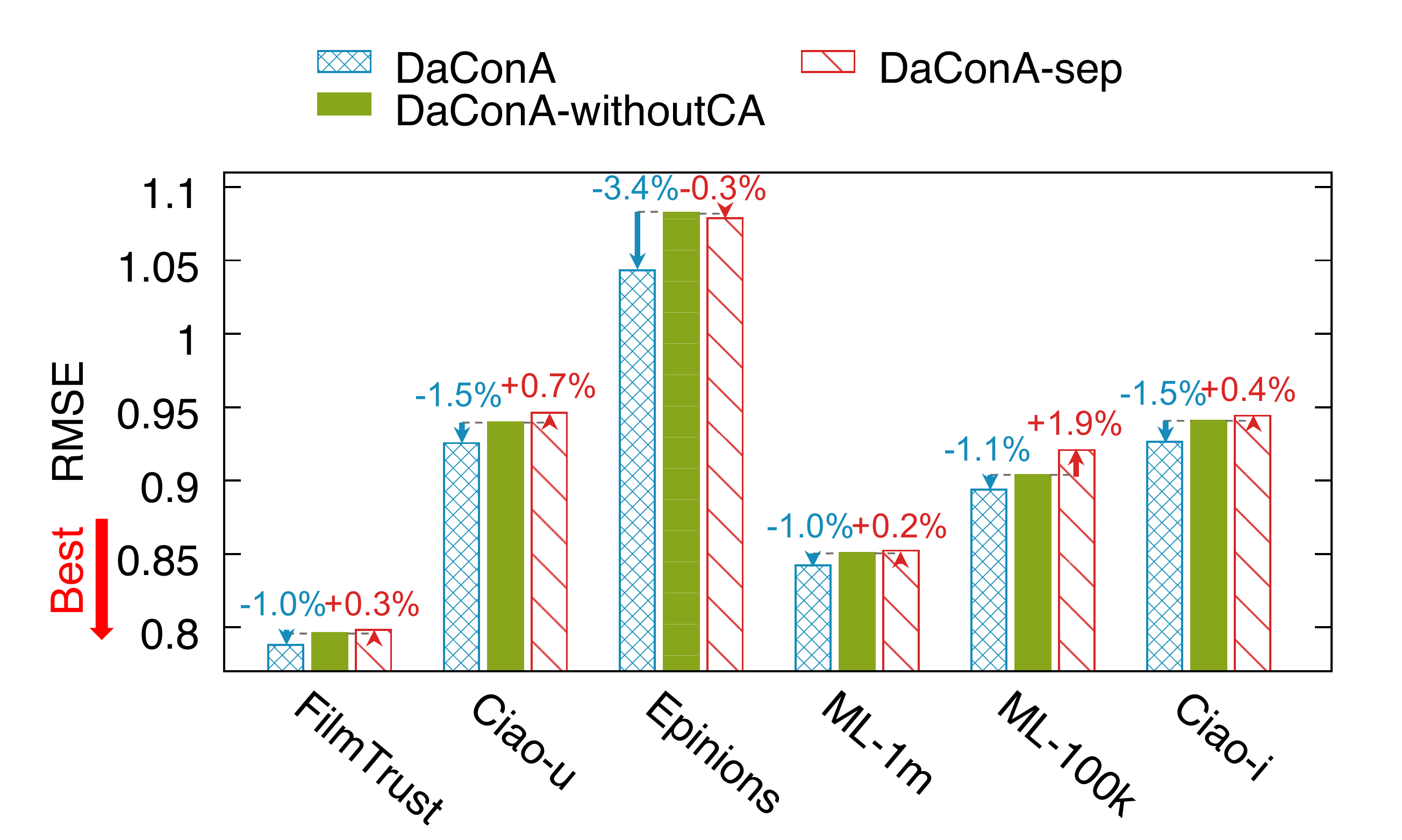}
	\vspace{-5px}
	\caption{
\method outperforms \method-$withoutCA$, showing that data context adaptation improves the accuracy.
\method also outperforms \method-$sep$, showing that sharing the same data context adaptation matrix improves the accuracy.
%
	}
	\label{fig:domain}
\end{figure}

We verify the effectiveness of data context adaptation layer in Figure~\ref{fig:domain}.
We compare three models: \method, \method-$withoutCA$, and \method-$sep$.
\method is our proposed method where we optimize data context adaptation matrices $D^X$ and $D^Y$ in training process.
\method-$withoutCA$ does not adapt to each data context, by setting $D^X$ and $D^Y$ to fixed (non-learnable) identity matrices.
To analyze the effect of setting a common adaptation matrix for each data context, we also devise \method-$sep$ where user and item have separate data context adaptation matrices for the rating data context (e.g., $D^{X-user}$ and $D^{X-item}$, respectively), and item and genre have separate data context adaptation matrices for the auxiliary data context (e.g., $D^{Y-item}$ and $D^{Y-genre}$, respectively).
For the three models, all hyperparameters are set to the optimal settings.
Figure~\ref{fig:domain} shows RMSEs of the methods in the real world datasets.
\method outperforms \method-$withoutCA$ for all datasets, which means that context adaptation layer is effective for the performance.
This is because \method projects the latent interaction vector through data context adaptation matrices that are trained to extract appropriate features for each data context, while \method-$withoutCA$ reuses the same latent interaction vector in different data contexts;
note that more performance improvements are observed for larger and sparser data (Epinions, Ciao-u, and Ciao-i).
Despite adding the learnable matrices, \method-$sep$ shows little change in performance when compared to \method-$withoutCA$; rather, the performances are degraded in almost all datasets (except for Epinions).
The results show that, in a data context, sharing the same data context adaptation matrix regardless of entities improves the generalization, while assigning different matrices per entity causes overfitting.


\subsection{Neural Networks (Q4)}
\label{subsec:450}
We investigate \method with different numbers of hidden layers to show whether deepening \method is beneficial to recommendation.
We also present the effectiveness of activation functions by comparing versions of \method with/without activation functions in terms of RMSE.

\begin{figure}[t]
	\centering
	\hspace{-15px}
	\includegraphics[width=0.42\textwidth]{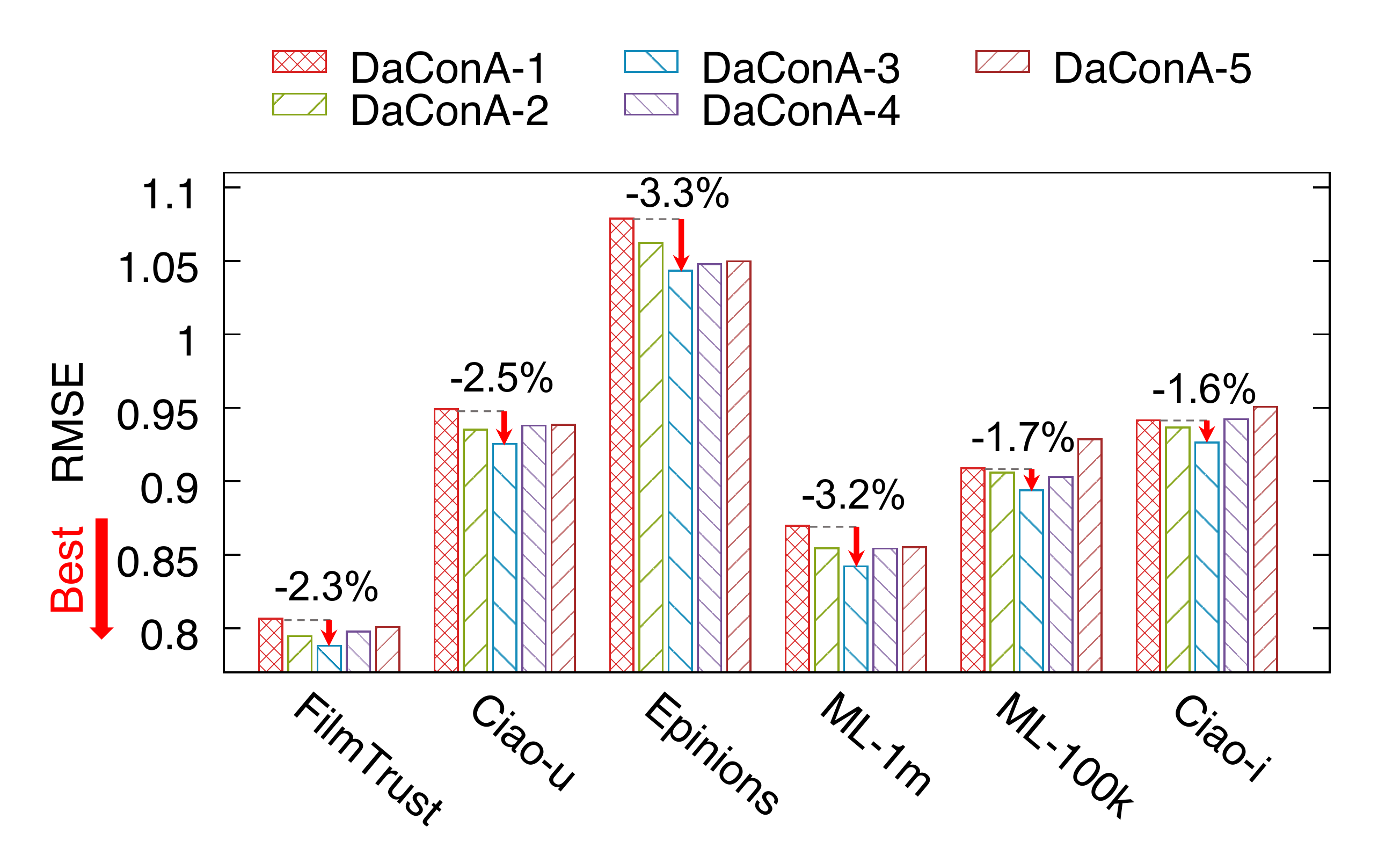}
	\vspace{-5px}
	\caption{
Using 3 layers instead of too small or too large layers gives the best accuracy.
\method-$k$ denotes \method with $k$ layers.
	Each percentage indicates the error reduction rate of the best model \method-$3$ compared to \method-$1$.
	}
	\label{fig:layers}
\end{figure}

First, we analyze whether deep layers of \method helps improve the performance in Figure~\ref{fig:layers}.
Since we set the depths of $f_X$ and $f_Y$ the same for all experiments, we define the depth of \method as the depth of $f_X$ or that of $f_Y$.
\method-$k$ denotes \method with depth $k$.
%
The structures of \method-$1$, \method-$2$, \method-$3$, \method-$4$, and \method-$5$ are [10], [20, 10], [40, 20, 10], [80, 40, 20, 10], and [160, 80, 40, 20, 10], respectively.
%
Figure~\ref{fig:layers} shows RMSEs of \method-$1$ to \method-$5$ on real-world datasets.
The results show that \method-$3$ has the best performance for all datasets.
Although increasing the number of layers from 1 to 3 helps improve the performance thanks to increased model capacity, further increasing the number of layers hurts the performance possibly due to overfitting.


%

\begin{figure}[t]
	\centering
	\hspace{-15px}
	\includegraphics[width=0.42\textwidth]{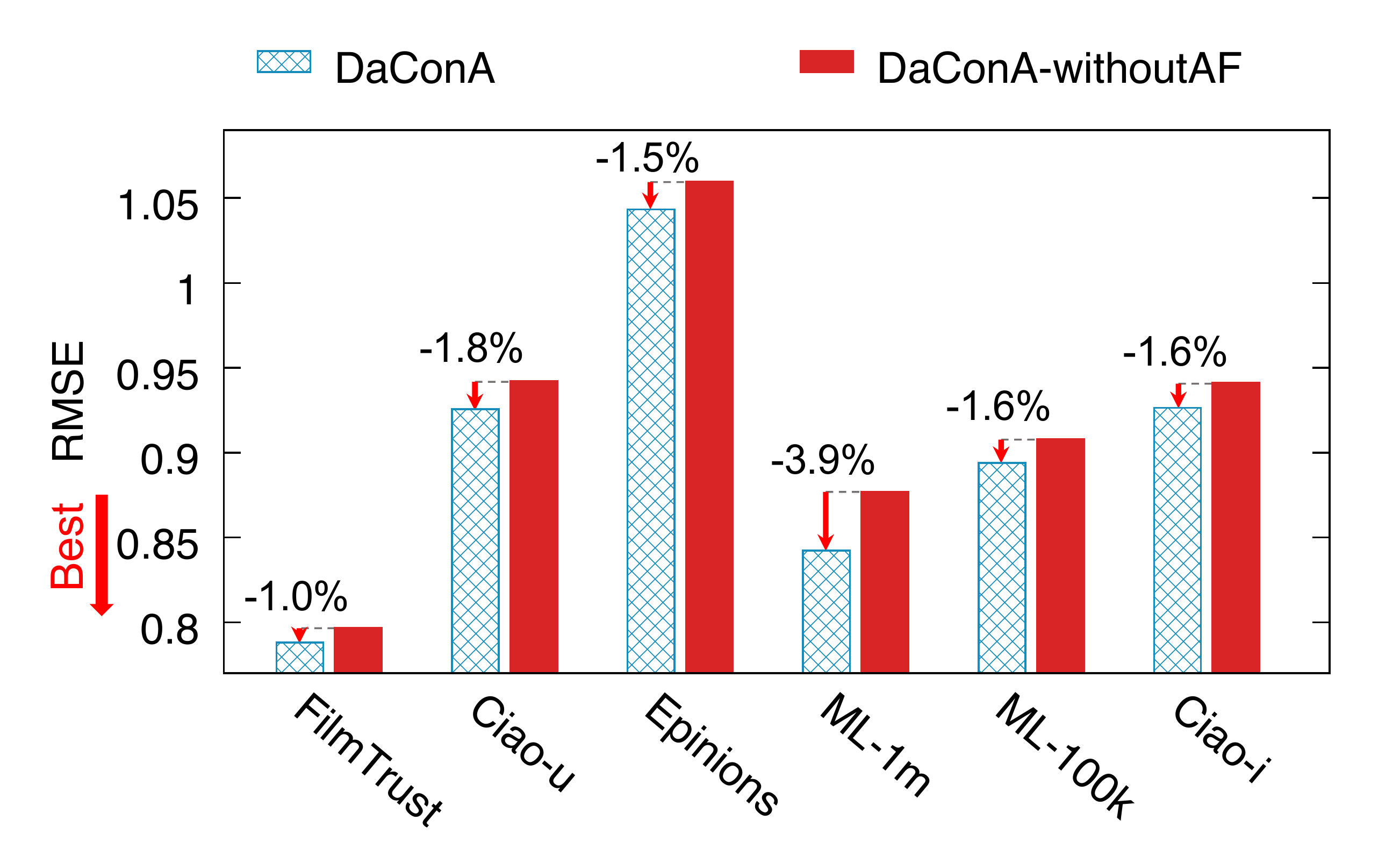}
	\vspace{-5px}
	\caption{
Non-linear activation function helps improve the accuracy in \method. 
%
	}
	\label{fig:activation}
\end{figure}

Second, we analyze whether non-linear activation functions of \method helps improve the performance in Figure~\ref{fig:activation}.
\method-$withoutAF$ denotes \method without non-linear activation functions.
The results show that the non-linear activation functions of \method improve the performance, which means activation functions are helpful in modeling non-linear relationships between entities in real-world datasets.
%
%


\section{Related Works}
\label{sec:related}


Many studies proposed methods that leverage additional information to alleviate the rating sparsity problem in collaborative filtering.
The methods are classified into two categories according to the type of additional data: user data context based methods, and item data context based methods.

\textbf{User data context.}
Studies \cite{chaney2015probabilistic, hao2008sorec, guo2015trustsvd, guo2014etaf, yang2012circle, ma2013experimental, qin2018dynamic} extensively proved that using not only rating data but also additional data about users mitigates rating sparsity problem in collaborative filtering.
Ma et al. \cite{hao2008sorec} proposed SoRec that uses rating matrix and truster-trustee matrix to predict unobserved ratings.
SoRec jointly optimizes latent item vectors using both rating matrix and trust matrix.
Yang et al. \cite{yang2012circle} proposed a method that extracts a subset of friends to be used in collaborative filtering.
Guo et al. \cite{guo2015trustsvd} proposed TrustSVD that combines user's implicit data and rating matrix based on biased matrix factorization \cite{koren2009mfrecommender}.

\textbf{Item data context.}
Previous works \cite{leung2006integrating, wang2011collaborative, wang2015collaborative, kim2016convolutional, hu2017integrating, bauman2017aspect, qin2018dynamic} proposed models that utilize item data context as additional information, and there also have been many methods using review data to provide personalized recommendations \cite{chen2015recommender}.
Leung et al. \cite{leung2006integrating} proposed a method that quantifies reviews through sentiment analysis and reflects it in rating prediction.
Wang et al. \cite{wang2011collaborative} combined topic modeling and collaborative filtering.
Wang et al. \cite{wang2015collaborative} integrated Stacked Denoising AutoEncoder (SDAE) ~\cite{vincent2010sdae} and Probabilistic Matrix Factorization (PMF) ~\cite{salakhutdinov2008pmf}.
Dong et al. \cite{dong2017hybrid} combined two Additional Stacked Denoising AutoEncoders (aSDAE) to support side information of both user and item.
Kim et al. \cite{kim2016convolutional, kim2017deep} proposed ConvMF that integrates Convolutional Neural Networks capturing contextual information of item documents into the PMF.
Hu et al. \cite{hu2017integrating} proposed a model that integrates item reviews into Matrix Factorization based Bayesian personalized ranking (BPR-MF).
Bauman et al. \cite{bauman2017aspect} proposed SULM that analyzes sentiment for each aspect by decomposing reviews into aspect units.
SULM predicts not only the probability that a user likes an item, but also which aspect will have a large effect.

\section{Conclusion}
\label{sec:conclusion}

We propose \method, a neural network based method that predicts missing rating values by exploiting rating matrix and auxiliary matrix.
\method overcomes three challenges: modeling effective embedding vectors to represent \textit{interaction} and \textit{independence} information, considering data context difference, and modeling non-linear relationship between entities.
We separate the embedding vectors into latent interaction vectors and rich latent independence vectors for entities.
\method uses data context adaptation layer to extract latent factors suitable for each data context.
Furthermore, \method uses neural networks to model complicated non-linear relationship between latent factors.
We show that \method is a generalized method which integrates many methods including MF, biased-MF, CMF, and biased-CMF as special cases.
Through extensive experiments we show that \method gives the state-of-the-art performance in recommendation with rating data and auxiliary matrix.
Future works include extending the method for a time-evolving setting. 

\bibliographystyle{IEEEtran}
\bibliography{paper}

\appendix
\section{Appendix}
\label{sec:appendix}

\subsection{Using Multiple Auxiliary Information}
\label{subsec:A10}
\textbf{Extension.}
\method supports using multiple auxiliary matrices.
Assume we are given an items' auxiliary matrix and a users' auxiliary matrix.
Let $X \in \mathbb{R}^{|\mathbb{U}| \times |\mathbb{I}|}$ be a rating matrix, $Y \in \mathbb{R}^{|\mathbb{I}| \times |\mathbb{C}|}$ be an items' auxiliary matrix, and $Z \in \mathbb{R}^{|\mathbb{U}| \times |\mathbb{T}|}$ be a users' auxiliary matrix.
In the following, let $Y$ denote a movie-genre matrix, and $Z$ denote a user-trustee matrix.
Then $\mathbb{U}, \mathbb{I}$, $\mathbb{C}$, and $\mathbb{T}$ indicate sets of users, items, genres, and trustees, respectively.
Then the objective function is extended as follows:
\begin{equation}\label{eq:totalloss_a}
\begin{aligned}
L = (1-\alpha-\beta)loss_X + \alpha loss_Y, + \beta loss_Z,
\end{aligned}
\end{equation}
where
$loss_X$ is Equation~(\ref{eq:lossX}), $loss_Y$ is Equation~(\ref{eq:lossY}), and 
\begin{equation}\label{eq:lossZ_a}
\footnotesize
\begin{aligned}
loss_Z &= \frac{1}{2} \sum_{(i, l) \in \Omega_Z} (\hat{Z}_{il} - Z_{il})^2 + \frac{\lambda}{2} Reg_Z.\\
\end{aligned}
\end{equation}
\noindent
$\Omega_Z$ contains observable entries in $Z$, $\hat{Z}_{il}$ is predicted entry, and $Reg_Z$ is a regularization term.
The predictive model for $Z$ is defined as follows: 
\begin{equation}\label{eq:predict_a}
\footnotesize
\begin{aligned}
\hat{Z}_{il} = f_Z
(
\begin{bmatrix}
	D^Z U_i \circ D^Z T_l \\ U^Z_i \\ T^Z_l \\
\end{bmatrix}
),
\end{aligned}
\vspace{-3px}
\end{equation}
\noindent
where $f_Z$ is a fully-connected neural network.
The predictive models for $X$ and $Y$ are defined in Equation~(\ref{eq:350}) (details in Section~\ref{subsec:350}). 
$T$ is the latent interaction matrix for the trustees, $U^Z$ is the latent independence matrix for the users in the user-trustee data context, and $T^Z$ is the latent independence matrix for the trustees in the user-trustee data context; note that the latent interaction matrix $U$ of users is trained from both $X$ and $Z$; similarly, $V$ is trained from both $X$ and $Y$.
Using this approach, it is trivial to further extend \method to utilize more than two auxiliary matrices (e.g., movie-director, movie-document, and user-demographic).

\begin{figure}[t]
	\centering
	\hspace{-10px}
	\includegraphics[width=0.3\textwidth]{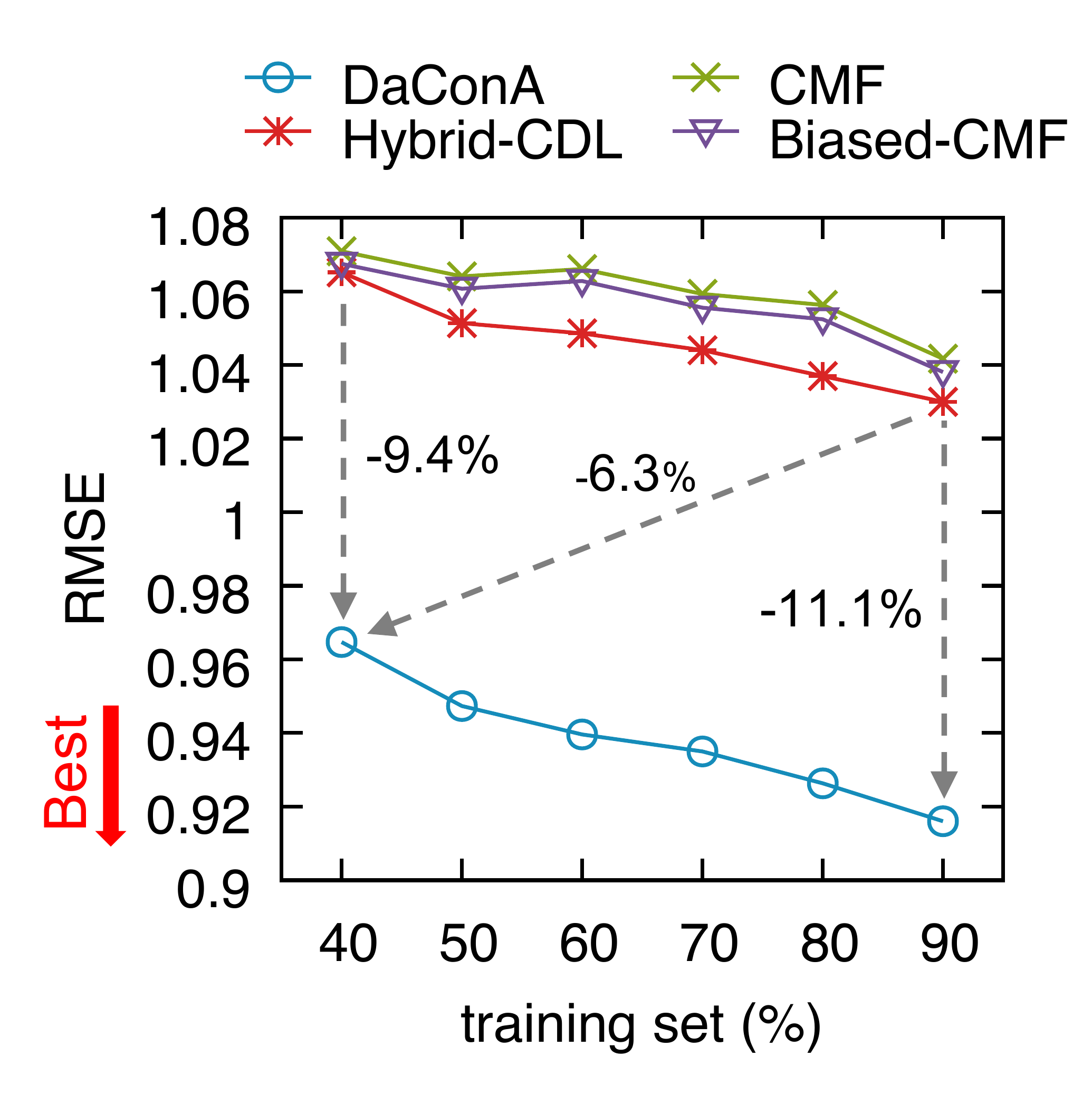}
	\caption{
\method gives the best accuracy even when using two auxiliary matrices. 
	Each arrow indicates the reduction rate of \method's RMSE compared to hybrid-CDL's RMSE.}
	\label{fig:ext}
\end{figure}

\textbf{Experiments.}
Among the competitors, CMF~\cite{singh2008relational}, Biased-CMF, and Hybrid-CDL~\cite{dong2017hybrid} which use multiple auxiliary information are chosen for detailed comparison;
even though FM also utilizes multiple auxiliary information, its poor performance clearly removes the need for detailed comparison (see Table~\ref{table:performance}).
We compare the models in Ciao\textsuperscript{\ref{ciao}} dataset which consists of a rating matrix and two auxiliary matrices: item-genre matrix and user-trustee matrix.
CMF and Biased-CMF are composed of three MFs and three Biased-MFs, respectively.
As described in Section~\ref{subsec:410}, Hybrid-CDL uses two Additional Stacked Denoising Autoencoders (aSDAE), one for users and the other for items.
We find the optimal hyperparameters by grid search.
For CMF and Biased-CMF, we set $l$ and $\lambda$ as 1e-4 and 1e-4, respectively.
For hybrid-CDL, we adopt [500, 10, 500]-structured hidden layers to the two aSDAEs, and set $l$, $\lambda$, $\alpha_1$, and $\alpha_2$ as 1e-3, 1e-3, 0.8, and 0.3, respectively; we also use corrupted inputs with a noise level of 0.3 as described in~\cite{dong2017hybrid}.
For \method, we use [40, 20, 10]-structured hidden layers, and set $l$, $\lambda$, $d_s$, $\alpha$, and $\beta$ as 1e-3, 1e-5, 11, 0.6, and 0.2, respectively.
Figure~\ref{fig:ext} shows the RMSEs while varying the percentage of training set; the rest of the data are used as test set.
Note that \method consistently outperforms the competitors by a significant margin, better utilizing auxiliary information.
Even \method using 40\% of training data shows 6.3\% lower RMSE than hybrid-CDL using 90\% of training data.

\end{document}